\pdfoutput=1

\documentclass[11pt]{article}

\usepackage[final]{acl}
\usepackage{xcolor} 
\usepackage{booktabs}  

\usepackage{times}
\usepackage{latexsym}
\usepackage{tcolorbox}
\usepackage{float}
\usepackage{placeins}
\usepackage{array}
\usepackage[T1]{fontenc}

\usepackage[utf8]{inputenc}

\usepackage{microtype}
\usepackage{todonotes}
\usepackage{booktabs}
\usepackage{inconsolata}
\usepackage[nolist ]{acronym}
\usepackage{comment}
\usepackage{graphicx}
\usepackage{amsmath} 

\usepackage{listings}

\begin{acronym}
\acro{ai}[AI]{Artificial Intelligence}
\acro{api}[API]{Application Programming Interface}
\acro{dq}[DQ]{Data Quality}
\acro{ml}[ML]{Machine Learning}
\acro{llm}[LLM]{Large Language Model}
\acro{lmm}[LMM]{Large Multimodal Model}
\acro{lpm}[LPM]{Large Process Model}
\acro{nlp}[NLP]{Natural Language Processing}
\acro{gpt}[GPT]{Generative Pretrained Transformer}
\acro{gptm}[GPT Model]{Generative Pretrained Transformer Model}
\acro{rnn}[RNN]{Recurrent Neural Network}
\acro{lstm}[LSTM]{Long Short-term Memory network}
\acro{gru}[GRU]{Gated Recurrent Unit}
\acro{bpm}[BPM]{Business Process Management}
\acrodef{bpmn}[BPMN]{Business Process Model and Notation}
\acro{ocr}[OCR]{Optical Character Recognition}
\acro{gpu}[GPU]{Graphical Processing Unit}
\acro{rag}[RAG]{Retrieval-Augmented Generation}
\acro{sft}[SFT]{Supervised Fine-Tuning}
\acro{hitl}[HITL]{Human-in-the-Loop}
\acro{rlhf}[RLHF]{Reinforcement Learning from Human Feedback}
\acro{dpo}[DPO]{Direct Preference Optimization}
\acro{peft}[PEFT]{Parameter-Efficient Fine-Tuning}
\acro{lora}[LoRA]{Low-rank adaptation}
\acro{qlora}[QLoRA]{Quantized Low-rank adaptation}
\acro{sota}[SOTA]{State-of-the-Art}
\acro{cot}[CoT]{Chain of Thought}
\acrodef{mmlu}[MMLU]{Massive Multi-task Language Understanding}
\end{acronym}

\definecolor{systemmsg}{RGB}{70,130,180}
\definecolor{usermsg}{RGB}{34,139,34}
\definecolor{assistantmsg}{RGB}{220,20,60}
\definecolor{codebg}{RGB}{248,248,248}

\newcommand{\rolelabel}[2]{%
    \textbf{\textcolor{#1}{#2:}}%
}

\title{(Towards) Scalable Reliable Automated Evaluation with Large Language Models}

\author{Bertil Braun \\
  KIT \\
  \texttt{bertil.braun@alumni.kit.edu} \\\And
  Martin Forell \\
  KIT \\
  \texttt{martin.forell@kit.edu} \\}

\begin{document}
\maketitle
\begin{abstract}
  Evaluating the quality and relevance of textual outputs from Large Language Models (LLMs) remains challenging and resource-intensive.
Existing automated metrics often fail to capture the complexity and variability inherent in LLM-generated outputs.
Moreover, these metrics typically rely on explicit reference standards, limiting their use mostly to domains with objective benchmarks.
This work introduces a novel evaluation framework designed to approximate expert-level assessments of LLM-generated content.
The proposed method employs pairwise comparisons of outputs by multiple LLMs, reducing biases from individual models.
An Elo rating system is used to generate stable and interpretable rankings.
Adjustable agreement thresholds—from full unanimity to majority voting—allow flexible control over evaluation confidence and coverage.
The method’s effectiveness is demonstrated through evaluating competency profiles extracted from scientific abstracts.
Preliminary results show that automatically derived rankings correlate well with expert judgments, significantly reducing the need for extensive human intervention.
By offering a scalable, consistent, and domain-agnostic evaluation layer, the framework supports more efficient and reliable quality assessments of LLM outputs across diverse applications.
\end{abstract}




\section{Introduction}

\acp{llm} are machine learning-based models capable of understanding, analyzing, and generating human language \cite{jarrahi_artificial_2023}.
Their advanced capabilities stem from extensive training on large-scale datasets, enabling them to develop a profound understanding of syntax, semantics, and contextual language aspects \cite{chang_survey_2024}.
Consequently, natural language processing has become a core component of \acp{llm}.
Recent advancements have significantly improved their capacity for semantic analysis and textual data comprehension \cite{deutsch_towards_2021,wu_large_2023}.
As a result, \acp{llm} are broadly employed across numerous domains, including software test generation \cite{schafer_empirical_2024}, question answering \cite{liang_holistic_2023}, and text summarization \cite{deutsch_towards_2021,pu_summarization_2023}.

Evaluating the quality of textual outputs generated by \acp{llm}, however, poses significant methodological challenges, primarily due to the inherently subjective and task-specific nature of text evaluation \cite{anwar_foundational_2024,chang_survey_2024}.
Traditional evaluation approaches typically depend on either human judgment—which is resource-intensive, inconsistent, and difficult to scale—or predefined metrics that are often insufficient to capture nuanced variations in quality across diverse tasks \cite{chiang_can_2023}.
These limitations highlight a critical gap in current evaluation methodologies, underscoring the necessity for more robust and scalable alternatives.

To address these evaluation challenges, this paper proposes a robust and scalable evaluation framework that leverages \acp{llm} themselves to perform systematic pairwise comparisons.
In contrast to conventional methods dependent solely on single-\ac{llm} judgments or fixed metrics, the presented approach integrates multiple \ac{llm} judgments and aggregates them using the Elo rating system.
This aggregation method produces reliable and consistent rankings, substantially reducing the need for extensive human evaluation.
Thus, the proposed method serves effectively as a universal evaluation layer applicable to a wide range of tasks involving free-form text generation.


The remainder of this paper is structured as follows: Section~\ref{sec:background} introduces foundational concepts, including \acp{llm}, the Elo rating system, and correlation metrics. Section~\ref{sec:related_work} provides an overview of related work.
Section~\ref{sec:approach} describes the proposed evaluation framework in detail, followed by a prototypical implementation in Section~\ref{sec:implementation}.
Section~\ref{sec:evaluation} demonstrates the framework’s applicability by evaluating its performance in extracting competency profiles from scientific abstracts and discusses the results.
Section~\ref{sec:conclusion} summarizes the main contributions and concludes the paper.
Finally, Section~\ref{sec:limitations} highlights the limitations of the proposed approach.



\section{Background} \label{sec:background}
This section briefly introduces foundational concepts of \acp{llm}, the Elo rating system, and correlation metrics, which are essential for understanding the proposed evaluation framework presented subsequently.
\subsection{Large Language Models}
\label{sec:large-language-models}

\acp{llm} have transformed \ac{nlp} through advanced machine learning methods, particularly the Transformer architecture, which efficiently captures long-range dependencies via self-attention mechanisms~\cite{vaswani_attention_2023}.
Modern \acp{llm}, such as GPT-4o~\cite{openai_gpt-4_2024}, Llama~3~\cite{metaai_llama_2024}, Mistral~\cite{jiang_mistral_2023}, and Phi~3~\cite{abdin_phi-3_2024}, represent the state of the art in diverse \ac{nlp} tasks, leveraging extensive pre-training on vast textual datasets.

To further enhance the quality and contextual appropriateness of outputs, various prompt engineering methods have emerged, notably \textit{Role Prompting} \cite{wang_role_2024}, \textit{Knowledge Injection} \cite{martino_knowledge_2023}, and \textit{\ac{cot}} \cite{wei_chain--thought_2023}.
Additionally, the \ac{rag} approach \cite{lewis_retrieval-augmented_2021} integrates retrieval mechanisms into text generation, allowing \acp{llm} to dynamically incorporate external domain-specific knowledge, thereby improving accuracy and relevance without extensive retraining.

\subsection{Elo Rating System for Ranking Items}
\label{sec:elo-rating-system}

The Elo rating system \cite{elo_rating_1986}, originally developed to rank chess players based on their relative skill levels, is a method for dynamically updating item rankings through pairwise comparisons.
Each item begins with an initial rating (e.g., 1000 points), which is adjusted after every comparison.

The Elo system uses the following formula to calculate the expected score for an item:
\[
    E = \frac{1}{1 + 10^{(\text{Rating}_{\text{opponent}} - \text{Rating}_{\text{player}})/400}}
\]
where \(E\) represents the expected probability of an item winning against its opponent. After a comparison, the rating is updated as:
\[
    \text{Rating}_{\text{new}} = \text{Rating}_{\text{current}} + K \times (\text{Score} - E)
\]
where \(K\) is a constant (typically 4 - 32) that determines the magnitude of rating adjustments, and \(\text{Score}\) is \(1.0\) for a win, \(0.0\) for a loss, and \(0.5\) for a draw.

By iterating this process across all pairwise outcomes, the Elo system produces a final ranked list of items. Items with consistently strong performance rise in rank, while those with frequent losses fall. This dynamic ranking approach ensures that the final rankings are both robust and reflective of the relative quality of the items.

\subsection{Correlation Metrics} \label{sec:correlation-metrics}
To assess agreement between automated evaluations and expert judgments, correlation metrics specifically suited for ordinal data are necessary. Spearman's rank correlation coefficient (Spearman’s $\rho$) measures the strength and direction of monotonic relationships between two ranked variables by comparing ranks rather than absolute values \cite{spearman_proof_2010}.
Kendall’s tau ($\tau$) similarly assesses rank correlation, but relies on pairwise comparisons, quantifying the proportion of concordant versus discordant rank pairs \cite{kendall_new_1938}.
Both metrics range from $-1$ to $+1$, where values near $+1$ indicate strong positive agreement, near $-1$ imply strong disagreement, and values close to $0$ suggest minimal or no correlation.
They do not assume linear relationships or normal distributions, making them particularly robust for evaluating ranked data in experimental settings.

\section{Related Work} \label{sec:related_work}
Evaluation of \acp{llm} has become increasingly crucial due to their widespread application. Reliable assessment methods are necessary to ensure outputs meet quality standards, motivating the development of various evaluation strategies. Existing methodologies typically fall into two categories: \textit{reference-based metrics} and \textit{reference-free methods}.

Reference-based metrics, such as \textit{BLEU} \cite{papineni_bleu_2002}, \textit{ROUGE} \cite{lin_rouge_2004}, and \textit{BERTScore} \cite{zhang_bertscore_2020}, assess outputs by comparing them to predefined reference texts.
However, their dependence on static references limits their applicability, especially for creative or open-ended tasks \cite{chang_survey_2024}.

To overcome this limitation, reference-free methods like \textit{GPTScore} \cite{fu_gptscore_2023} have emerged, directly evaluating outputs based on token probabilities and task-specific dimensions.
Although these approaches are promising, they sometimes exhibit limited correlation with human judgments~\cite{fu_gptscore_2023}, highlighting the ongoing need for more accurate evaluation techniques.

An alternative approach, known as \textit{LLM-as-a-Judge} \cite{zheng_judging_2023}, utilizes \acp{llm} themselves to perform evaluations.
This can be implemented either through single-\ac{llm} scoring or through more robust multi-\ac{llm} frameworks, such as debates or peer reviews~\cite{chang_survey_2024, liang_holistic_2023}.

Within multi-\ac{llm} evaluation frameworks, the Elo rating system has gained popularity as a structured method for dynamically ranking models based on pairwise comparisons. Despite its widespread use, Elo ratings are sensitive to factors such as evaluation order and hyperparameter selection, leading to reliability concerns~\cite{boubdir_elo_2023}. Recent work by~\cite{boubdir_elo_2023} proposes guidelines to enhance reliability, including a permutation oversampling approach to mitigate order effects, thereby enabling a more robust and dependable model performance assessment.

\section{Approach} \label{sec:approach}

\begin{figure}[ht]
    \centering
    \includegraphics[width=\linewidth]{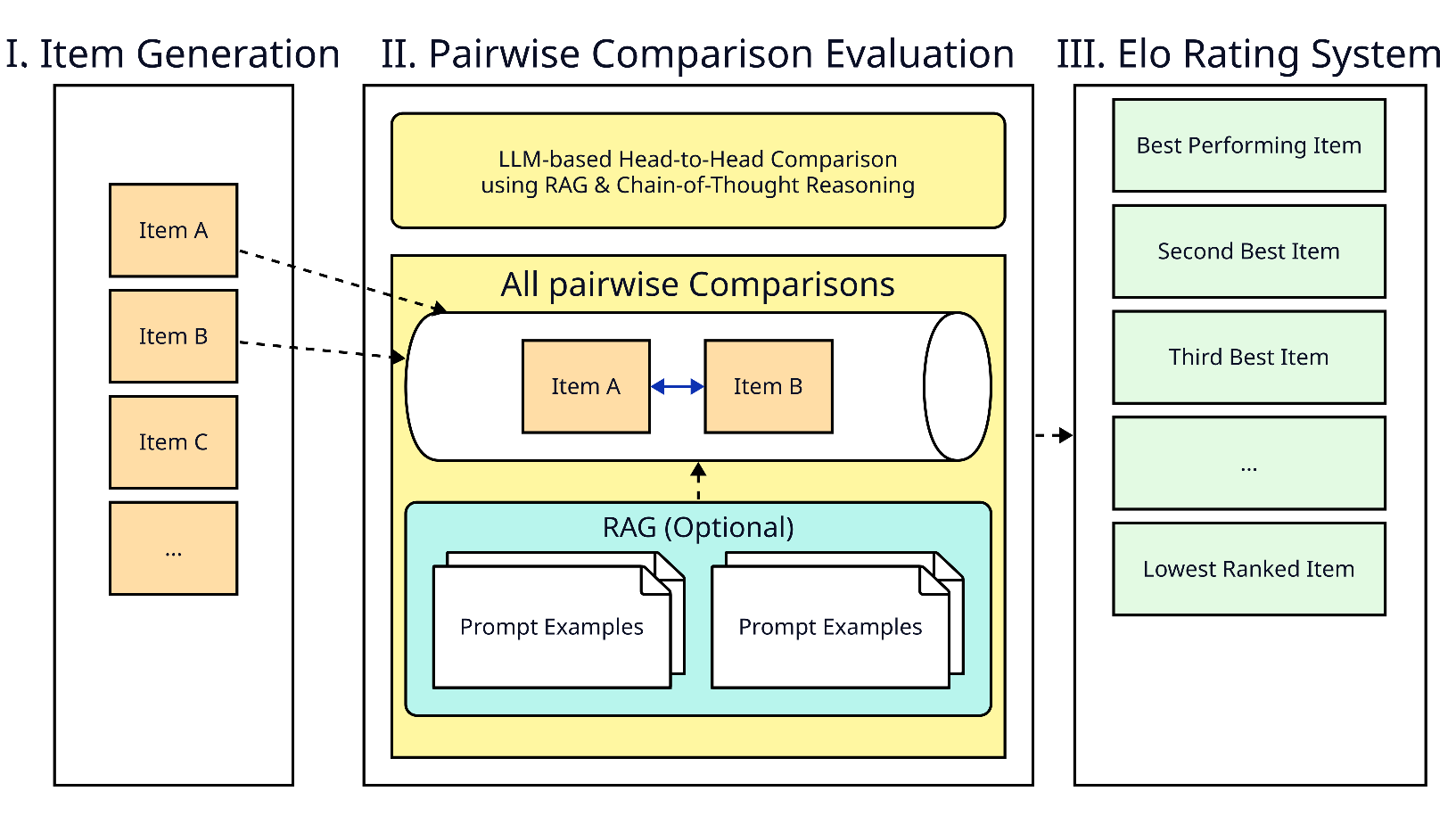}
    \caption{Pipeline Overview: A three-stage methodology including Generation, Comparison, and Ranking.}
    \label{fig:pipeline-overview}
\end{figure}

\noindent\textit{Overview and Motivation.}
This section presents a methodology for utilizing \acp{llm} to assess diverse free-text responses to a given task (e.g., summarization) through a pairwise comparison methodology. Evaluating free-text outputs with \acp{llm} poses several challenges:
\begin{itemize}
    \item \textit{C1 – Subjectivity in Scoring:} Absolute scores are often inconsistent and subject to scaling issues.
    \item \textit{C2 – LLM Biases:} Positional, verbosity, and stylistic biases can distort evaluation outcomes.
    \item \textit{C3 – Handling Multiple Evaluations:} Aggregating multiple LLM outputs into a coherent decision is non-trivial.
    \item \textit{C4 – Robust Ranking:} Deriving a definitive ordering of items in a bias-minimized fashion requires a resilient aggregation mechanism.
\end{itemize}
To address these challenges, our pipeline is organized into three distinct stages: (I) generation of items to compare, (II) systematic pairwise comparison using multiple \acp{llm}, and (III) ranking the items with an Elo rating system to clearly identify the best-performing candidates. Figure~\ref{fig:pipeline-overview} outlines this pipeline.

The methodology begins with generating multiple items intended for comparison.
This step may include various methodologies to ensure diverse inputs for evaluation.
For example, different hyperparameter configurations, distinct \acp{llm}, or alternative wording styles can be employed.
The methodology then systematically assesses and ranks the items, enabling the identification of the methodology with the best results for the given task.

Typical applications include hyperparameter optimization, method comparison, and \ac{llm} selection, where the objective is to determine the most effective configuration or \ac{llm}.

Based on the final Elo ranking, the performance of different methods is assessed, and the best-performing item is identified.
This highest-ranked item can subsequently be deployed in production environments or research settings.

\subsection{Pairwise Comparison Framework}

\begin{figure}[ht]
    \centering
    \includegraphics[width=\linewidth]{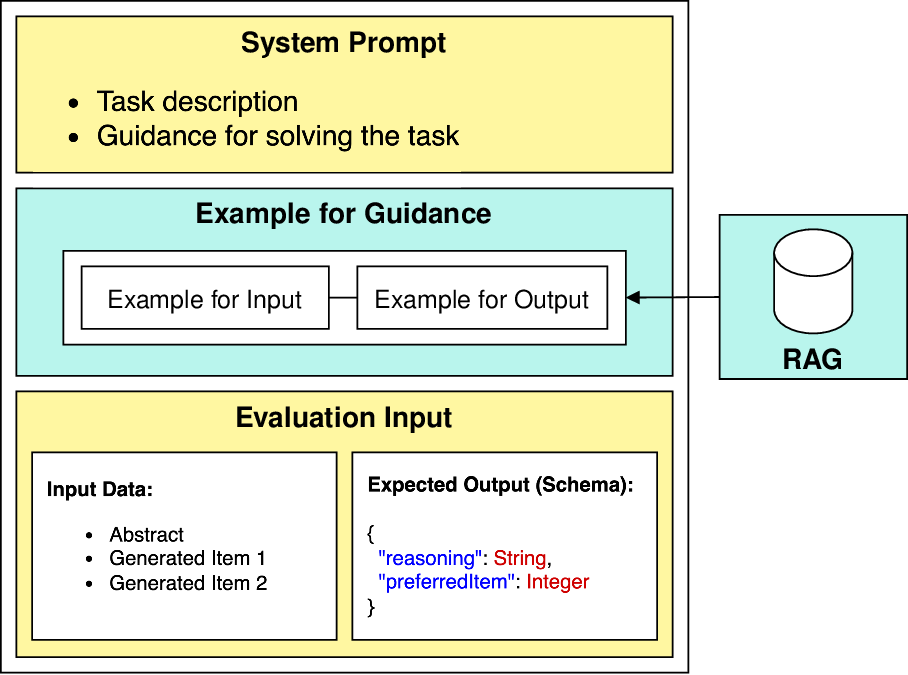}
    \caption{Overview of the message prompt used in the evaluation methodology.}
    \label{fig:prompt}
\end{figure}

\noindent\textit{(Addresses C1 – Subjectivity in Scoring)}
The evaluation methodology builds upon a pairwise comparison methodology designed to deliver precise and consistent evaluations. Instead of assigning absolute scores—which are susceptible to subjectivity and scaling inconsistencies \cite{liu_aligning_2025, gu_survey_2025}—the focus lies on relative judgments through direct item-to-item comparisons.
Two items are presented simultaneously to an \ac{llm}, with evaluation criteria explicitly defined by the user based on the specific task. For instance, in summarization tasks, the criterion might be: "which item better summarizes {a given text}?".

The methodology incorporates established \ac{sota} prompting techniques, including Role Prompting, \ac{cot}, and Self-Consistency Decoding, to optimize performance, enhance consistency, and mitigate biases from \acp{llm}. The evaluation methodology leverages \ac{rag} by embedding contextually relevant examples directly into the prompt, effectively creating a few-shot scenario \cite{brown_language_2020}.
These examples illustrate appropriate evaluation practices, providing clear task demonstrations (see Figure~\ref{fig:prompt}). A complete example of the prompt structure used in our evaluation framework is provided in Appendix~\ref{example:prompt}.
The prompt requires the \ac{llm} to engage in a chain-of-thought process, articulating its reasoning step-by-step before deciding on the item that best aligns with the specified criteria.
Additionally, a system prompt ensures a structured output format, systematically presenting detailed reasoning alongside the final decision.
This integrated strategy ensures systematic, transparent, and goal-aligned evaluations, supporting robust downstream analyses.

All pairwise comparison outcomes are fully automated. Once the prompts and task-specific evaluation criteria are defined, no human judgment is involved in determining which item wins a given comparison. Each \ac{llm} receives the same structured prompt with fixed instructions and examples, and the final rankings are derived solely from the aggregated Elo updates based on these model judgments.

Given \(n\) items, the total number of unique comparisons is \(\frac{n \times (n - 1)}{2}\).
To mitigate positional biases, each pair is evaluated bidirectionally — posing both questions: "Is A better than B?" and "Is B better than A?" to the \acp{llm}.
This strategy ensures that evaluation results remain independent of item presentation order.
Additionally, multiple \acp{llm} are utilized for each comparison, significantly enhancing the robustness of the methodology.
For \(n\) items evaluated by \(N_{\mathrm{LLM}}\) \acp{llm}, the total number of pairwise evaluations is \(n \times (n - 1) \times N_{\mathrm{LLM}}\).

To ensure consistency, all \acp{llm} receive identical prompts, and the evaluation criteria remain fixed.

\subsection{Mitigation of LLM Biases}

\noindent\textit{(Addresses C2 – LLM Biases)}
\acp{llm} exhibit various biases that can compromise the reliability of evaluations. Positional bias is one prominent issue, with \acp{llm} often favoring the last-presented option in pairwise comparisons, as highlighted by \cite{zhao_survey_2025}.
Additionally, verbosity bias, which favors longer or more elaborate responses regardless of quality, is common \cite{zhao_calibrate_2021}.
Stylistic biases, including preferences for particular syntactic structures or formality, also potentially skew evaluations involving language variation \cite{lewkowycz_solving_2022}.
If unaddressed, these biases can introduce systematic errors into evaluations.

To mitigate these biases, several strategies are incorporated:

First, prompts are meticulously crafted using neutral and unbiased language to avoid unintentionally influencing the \ac{llm}’s judgment. Furthermore, prompt consistency across evaluations minimizes variability arising from prompt design.

Second, bidirectional evaluations counter positional bias by reversing the presentation order of items in comparisons, thereby reducing order-induced preferences.

Third, \ac{rag} techniques are utilized. Given a database containing relevant examples from previous evaluations (for example, expert-reviewed domain-specific comparisons), the most contextually similar example is retrieved and included in the prompt. This provides the \ac{llm} with concrete demonstrations of previously applied criteria in similar contexts, enabling more informed, criteria-consistent evaluations and reducing potential biases through recognizing patterns and contextual commonalities.

\subsection{Handling Multiple Evaluations and Agreement Thresholds}
\label{sec:agreements}

\noindent\textit{(Addresses C3 – Handling Multiple Evaluations)}
When multiple \acp{llm} are used to evaluate item pairs, it is necessary to reconcile potentially differing judgments in a principled way. We explore two main strategies for aggregating multiple outputs: (1) threshold-based consensus, and (2) individual updates without aggregation.

In the threshold-based setting, an agreement threshold is specified (e.g., 100\%, 90\%, 75\%, 50\%) that determines whether a pairwise judgment results in a win/loss or a draw. These thresholds correspond to intuitive decision modes: \emph{consensus} (1.0), \emph{near-consensus} (0.9), \emph{qualified majority} (0.75), and \emph{simple majority} (0.5). If the specified threshold is met—for example, at least 75\% of \acp{llm} agree that item A is better than item B—an Elo update is performed accordingly. Otherwise, the comparison is treated as a draw, resulting in no net change in Elo scores. Higher thresholds prioritize certainty but produce more draws and may limit informativeness; thresholds below 0.5, in contrast, allow minority judgments to dominate and are typically avoided.

An alternative approach is to treat each \ac{llm}'s judgment independently, applying Elo updates for each evaluation. Rather than collapsing multiple judgments into a single binary decision, this method—termed the \emph{No-Threshold} variant—aggregates signal proportionally. For example, if 80\% of \acp{llm} prefer A over B, the cumulative updates represent a net 60\% push in favor of A, without discarding minority votes or reducing the result to a draw. This method retains more of the available information and avoids the over-conservatism introduced by strict agreement requirements.

\subsection{Elo Rating System for Ranking Items}

\noindent\textit{(Addresses C4 – Robust Ranking)}
Following the completion of pairwise comparisons, the results are aggregated into a global ranking using an Elo rating system (see \ref{sec:elo-rating-system}), producing a definitive ordered list of items. The Elo system is particularly suitable due to its dynamic updating mechanism based on pairwise outcomes. Items with consistently positive outcomes improve in ranking, while those frequently losing decline. This iterative method ensures final rankings robustly and accurately reflect the relative quality of the evaluated items. In addition, we hereby adhere to the guidelines proposed by \cite{boubdir_elo_2023}, by sampling multiple permutations of the LLM evaluations and applying the Elo rating system to each permutation. This approach ensures that the final ranking is not overly influenced by any single evaluation order, enhancing robustness and reliability.

\paragraph{Interpretability of Elo scores.}
A convenient property of Elo is that a score \emph{difference} ($\Delta$) maps directly to an expected win-probability via
\[
    P(\text{A beats B})=\frac{1}{1+10^{-\Delta/400}}.
\]
For example, $\Delta=100$ implies that item~A should win about $64\,\%$ of head-to-head comparisons with item~B, whereas $\Delta=200$ raises that expectation to roughly $76\,\%$.
We therefore encourage practitioners to report not only the final rank ordering but also the Elo gaps between adjacent candidates.
A task-agnostic rule-of-thumb is:
\begin{itemize}
  \item $\Delta<50$\,pts — items are practically tied;
  \item $50\le\Delta\le150$\,pts — a noticeable but moderate quality gap;
  \item $\Delta>150$\,pts — a strong, user-perceivable difference.
\end{itemize}
Publishing these gaps alongside ranks helps downstream readers understand \emph{how much better} one output is expected to be, not merely \emph{which} one is on top.

\section{Implementation} \label{sec:implementation}

The proposed evaluation pipeline has been implemented and is demonstrated through a specific use case: generating competency profiles from research abstracts (see Section~\ref{sec:competency-profile-definition}).
This scenario illustrates how the framework can be applied to real-world data and highlights its effectiveness in evaluating complex, task-specific outputs.
Competency profiles serve as a concrete example of evaluable items throughout the following sections.
The implementation leverages widely adopted tools and frameworks to ensure scalability, usability, and reliability.
This section details the technical stack, the integration of \acp{llm}, data sources, and the experimental setup. Additionally, it discusses implementation challenges and the strategies used to address them.

\subsection{Structured Competency Profiles}
\label{sec:competency-profile-definition}

A competency profile is defined as a structured summary of the research capabilities demonstrated by the authors of a given set of academic papers. Specifically, it identifies the overarching research domain in which the authors operate, alongside a set of 5 to 8 competencies that reflect key areas of expertise. Each competency is accompanied by a brief description (1–2 sentences) outlining its scope and relevance (see Appendix \ref{example:competency-profiles} for examples). To generate such profiles, a \ac{llm} is prompted with the abstracts of the input papers and tasked with inferring both the general domain and the detailed competencies exhibited across the works.

To evaluate the accuracy of these generated profiles, the evaluation \acp{llm} are provided with the same set of paper abstracts and asked to assess the extent to which each profile aligns with the actual competencies evidenced in the papers. This comparative evaluation focuses on the fidelity and relevance of the proposed domain and competencies relative to the source material.

\subsection{Integration of Large Language Models}
\label{sec:multi-llm}

The pipeline incorporates multiple \ac{sota}-\acp{llm}, selected based on their diverse capabilities and strong performance across a range of tasks (see Appendix~\ref{sec:llms-used}).
Access is provided through the free-tier or low-cost \acp{api} offered by platforms such as GROQ\footnote{\url{https://groq.com/}, Accessed: 2025-04-10}, OpenAI\footnote{\url{https://openai.com/}, Accessed: 2025-04-10}, and Google AI\footnote{\url{https://ai.google.dev/}, Accessed: 2025-04-10}, enabling broad experimentation and scalability.
Each \ac{llm} delivers robust text generation and comparison capabilities, aligning with the demands of both competency profile generation and pairwise evaluation.
Although proprietary constraints (e.g., details regarding quantization or other internal optimizations) remain undisclosed, they do not hinder the effective application of these \acp{llm} within the pipeline.

In the pipeline, the \emph{llama-3.1-70B} \cite{team_meta-llamallama-31-70b-instruct_2024} \ac{llm} generates competency profiles from research abstracts, employing a higher temperature setting and multiple completions (six per abstract) to enhance diversity and comprehensiveness of outputs.
Subsequently, \acp{llm} including \emph{gemma2-9b-it} \cite{gemma_2024}, \emph{llama-3.1-8b} \cite{team_meta-llamallama-31-8b_2024}, \emph{gpt-4o-mini} \cite{team_gpt-4o_2024}, \emph{gemini-2.0-flash} \cite{team_gemini_2025}, and \emph{mixtral-8x7b} \cite{team_mistralaimixtral-8x7b-instruct-v01_2024} perform pairwise evaluations of these generated profiles according to the previously established evaluation pipeline. This combined use of multiple models enhances robustness and reduces potential biases associated with relying on a single \ac{llm}.

\subsection{Data Sources}

The primary input data for competency profile generation is derived from research publications and their abstracts.
Abstracts are obtained from publicly accessible repositories such as the \textit{KITopen}\footnote{\url{https://www.bibliothek.kit.edu/kitopen.php}, Accessed: 2025-04-10} and \textit{OpenAlex}\footnote{\url{https://openalex.org/}, Accessed: 2025-04-10}.
To preserve the integrity of the data, minimal preprocessing is performed; the raw abstracts are passed directly to the \acp{llm}, ensuring authenticity and consistency in evaluation.

\subsection{Implementation Challenges and Solutions}

While the implementation was largely straightforward due to the availability of established tools and \acp{api}, certain challenges were encountered:

\paragraph{Scalability} Handling the large number of \ac{api} requests required for pairwise evaluations across multiple \acp{llm} posed a potential bottleneck. This was addressed by implementing efficient request handling and parallelization, ensuring that evaluations could scale with the size of the dataset.

\paragraph{Contextual Consistency} The \ac{llm} consistency initially exhibited significant inconsistency; Applying \ac{sota} prompting techniques, including \ac{rag}, chain-of-thought reasoning, and structured outputs, substantially improved inter-model and intra-model consistency across repeated evaluations, without any manual correction or human-in-the-loop tuning.

\section{Evaluation} \label{sec:evaluation}

To evaluate the proposed method for automated evaluation using \acp{llm}, an experimental study was conducted. This section outlines the evaluation strategy, introduces the dataset used, and presents the metrics and results related to \ac{llm} quality.
\subsection{Evaluation Strategy}
\label{sec:evaluation-strategy}

The evaluation strategy is based on an experimental setup that compares automated rankings generated by multiple \acp{llm} with expert judgments.
A total of 20 experts participated, each selecting 5--10 of their own publicly available research publications.
Experts initiated the process themselves by providing the abstracts of these publications, which ensured that any shared materials were already in the public domain.
Only abstracts were used in the experiments, thereby omitting personal identifiers such as author names or affiliations.
Although the content of the abstracts could theoretically allow an individual expert to be identified, no sensitive personal information was collected or processed in this study.

The selected abstracts were processed by various \acp{llm} to generate competency profiles.
The resulting profiles were evaluated using two distinct ranking methods:
(1) \textit{manual expert rankings}, wherein participants assessed the quality and relevance of the generated profiles in relation to their actual expertise via a web interface,
and (2) \textit{automated rankings}, produced through an Elo rating pipeline that aggregated pairwise comparisons performed by the \acp{llm}.

To assess the alignment between automated and expert-generated rankings, correlation-based metrics as described in Section~\ref{sec:metrics} were applied.
In addition, an ablation study using a single \ac{llm} was conducted to explicitly illustrate the impact of combining multiple \acp{llm}.


\subsection{Evaluation Metrics}
\label{sec:metrics}

To quantify the degree of agreement between automated and expert-generated rankings, the correlation metrics introduced in Section~\ref{sec:correlation-metrics} are applied: Spearman's rank correlation coefficient (Spearman’s \(\rho\)) and Kendall's tau (\(\tau\)).
These metrics are particularly appropriate for ordinal ranking comparisons, effectively capturing both monotonic relationships and pairwise rank agreement without relying on assumptions of linearity or normality.

\subsection{Results and Analysis}

We evaluate two primary strategies for integrating multiple \ac{llm} evaluations into an Elo-based ranking: Threshold-Based Consensus and No Threshold updates as described in Section~\ref{sec:agreements}. We first present results from a multi-\ac{llm} setup that pools judgments across all available \acp{llm}, followed by a single-\ac{llm} analysis using \texttt{llama-3.1-8b}. Spearman’s $\rho$ and Kendall’s $\tau$ correlations with expert rankings are reported alongside standard deviations and p-values.

\begin{table*}[hbt!]
    \centering
    \caption{Correlation between Elo-based and expert rankings with all \acp{llm}. "No T." indicates the No Threshold approach where every \ac{llm}'s judgment triggers an update.}
    \label{tab:multi-model-results}
    \begin{tabular}{l c c c}
        \toprule
        \textbf{Threshold} & \textbf{Spearman}          & \textbf{Kendall}           & \textbf{P-Value (Spearman / Kendall)} \\
        \midrule
        1.0                & 0.650 $\pm$ 0.211          & 0.560 $\pm$ 0.196          & 0.259 / 0.322                         \\
        0.9                & 0.660 $\pm$ 0.224          & 0.580 $\pm$ 0.227          & 0.256 / 0.315                         \\
        0.75               & 0.770 $\pm$ 0.219          & 0.720 $\pm$ 0.223          & 0.165 / 0.188                         \\
        0.5                & \textbf{0.830 $\pm$ 0.190} & \textbf{0.780 $\pm$ 0.209} & \textbf{0.114 / 0.142}                \\
        No T.              & 0.820 $\pm$ 0.183          & 0.760 $\pm$ 0.196          & 0.118 / 0.148                         \\
        \bottomrule
    \end{tabular}
\end{table*}
\begin{table*}[hbt!]
    \centering
    \caption{Correlation between Elo-based and expert rankings using only \texttt{llama-3.1-8b}. “No T.” is the No Threshold approach.}
    \label{tab:single-model-results}
    \begin{tabular}{l c c c}
        \toprule
        \textbf{Threshold} & \textbf{Spearman}          & \textbf{Kendall}           & \textbf{P-Value (Spearman / Kendall)} \\
        \midrule
        1.0                & 0.730 $\pm$ 0.224          & 0.660 $\pm$ 0.237          & 0.196 / 0.243                         \\
        0.9                & 0.760 $\pm$ 0.196          & 0.660 $\pm$ 0.220          & 0.162 / 0.235                         \\
        0.75               & 0.740 $\pm$ 0.291          & 0.560 $\pm$ 0.564          & 0.202 / 0.265                         \\
        0.5                & \textbf{0.850 $\pm$ 0.201} & \textbf{0.780 $\pm$ 0.227} & \textbf{0.100 / 0.152}                \\
        No T.              & 0.750 $\pm$ 0.206          & 0.660 $\pm$ 0.220          & 0.173 / 0.235                         \\
        \bottomrule
    \end{tabular}
\end{table*}
\subsubsection{Multi-Model Results}

Table~\ref{tab:multi-model-results} shows that very high thresholds (1.0, 0.9) yield moderate correlations but suffer from a high draw rate, since even minimal disagreement nullifies a comparison. Lowering the threshold to 0.75 captures more partial agreements and improves performance substantially. A simple majority requirement (0.5) provides the best average correlations, and using No Threshold (“No T.” in the table) is similarly effective. Notably, the modest difference between 0.5 and No Threshold suggests that Elo readily absorbs and balances minor disagreements when multiple \acp{llm} are involved.

\subsubsection{Single-Model Results}

Table~\ref{tab:single-model-results} illustrates that a single \ac{llm}, here \texttt{llama-3.1-8b}, does not benefit from cross-\ac{llm} disagreement in the same way.
While relaxing the threshold to 0.5 again delivers the strongest correlations, the No Threshold approach drops in effectiveness: contradictory judgments cannot be offset by another \ac{llm}'s consensus.
Consequently, No Threshold ranks below 0.5 in this scenario, even though both outpace higher thresholds such as 0.9 and 1.0.

\subsubsection{Observations and Takeaways}

Overall, requiring strong consensus (e.g., 90\% or 100\%) frequently introduces too many draws and discards partial-but-informative judgments, resulting in weaker correlations with expert rankings.


Loosening the threshold to a simple majority (0.5) allows more comparisons to produce decisive wins or losses, clearly boosting Elo performance. In the multi-\ac{llm} case, even the fully inclusive No Threshold option works well, suggesting that diverse \acp{llm} collectively moderate each other’s noise. However, in a single-\ac{llm} context, No Threshold tends to admit contradictory signals that are not corrected by other \acp{llm}, which slightly reduces ranking accuracy compared to a 0.5 threshold. These findings indicate that draws should not be overused, and that leveraging every moderate agreement signal is beneficial—particularly when multiple \acp{llm} are available to balance noise.

On average, adjacent ranks differed by $107$ Elo points when a consensus threshold ($1.0$–$0.50$) was used and by $159$ points under the No-Threshold setting.

Correlations with expert rankings remain stable—within ${\pm}0.03$—when varying the threshold between $0.50$ and $0.75$, indicating that final rankings are robust to this choice.




\section{Conclusion} \label{sec:conclusion}

This paper presented a scalable and reliable automated evaluation framework utilizing multiple \acp{llm} in combination with an Elo rating system, significantly enhancing the efficiency and consistency of assessments of \ac{llm}-generated texts.
The conducted evaluation demonstrated a strong alignment between automated rankings and expert judgments, thereby validating the multi-\ac{llm} approach.
The versatility of the presented framework supports broad applicability across diverse domains requiring nuanced textual evaluation, substantially reducing dependency on extensive human intervention.
Further research is encouraged, particularly focusing on optimizing computational efficiency to fully leverage the framework’s potential at scale.

\section{Limitations}\label{sec:limitations}
A primary limitation of our approach stems from the substantial computational overhead associated with inference-heavy pairwise comparisons. Specifically, the Elo-based evaluation requires \(\mathcal{O}(n^2)\) comparisons, each necessitating multiple LLM inferences, including bidirectional checks. This quickly becomes computationally intensive and potentially costly when employing large commercial \acp{llm}, even for moderately sized evaluation sets.

To mitigate the computational complexity, future work could investigate comparison-based sorting algorithms, aiming to reduce the required number of evaluations from \(\mathcal{O}(n^2)\) down to \(\mathcal{O}(n \log n)\) or even \(\mathcal{O}(n)\). Preliminary attempts at such sorting methods have encountered challenges, including frequent draws and a lack of guaranteed transitivity in comparisons produced by \acp{llm}. Nevertheless, Elo ratings currently provide a stable numeric metric, highlighting closely matched profiles and indicating areas of uncertainty effectively.

\label{sec:too-similar}
Another practical challenge arises when comparing highly similar items. When the differences between items are subtle, individual \ac{llm} evaluations can yield divergent outcomes due to inherent model biases and the varying strengths and weaknesses across different models. This variability complicates the task of reliably distinguishing between items of near-equivalent quality and can reduce the clarity and interpretability of rank-based evaluation methods.

In such cases, the Elo rating provides valuable insight into the relative quality of items, even when absolute differences are minimal. In several instances, items with only marginal quality distinctions received nearly identical Elo scores—an outcome that is informative in its own right. Notably, Elo-based rankings also help surface atypical items—either exceptionally strong or weak—when their scores deviate significantly from the rest, offering a robust signal for identifying outliers within a set of closely matched candidates.

Finally, the scope and robustness of our study remain constrained by the current size and diversity of the expert pool. Although the initial correlations observed between automated and expert rankings are promising, expanding the evaluation across a broader spectrum of academic disciplines and increasing the sample size through ongoing expert recruitment would significantly enhance the validity and generalizability of the presented results.

\section*{Acknowledgments}\label{sec:acknowledgments}

We acknowledge the use of \ac{ai} tools in this work.
Specifically, ChatGPT (\url{https://chat.openai.com/}) was employed for language assistance (i.e., paraphrasing, polishing, and proofreading), and GitHub Copilot (\url{https://github.com/features/copilot}) provided programming support.
Additionally, generative \ac{ai} assisted in identifying relevant evaluation metrics during the literature search.
All authors remain fully responsible for the accuracy and originality of the methods, results, and conclusions presented in this paper.

\section*{ETHICAL IMPACT STATEMENT}\label{sec:ethical-impact}

\paragraph{Adherence to the ACL Code of Ethics.}
Our work aligns with the principles outlined in the ACL Code of Ethics (\url{https://www.aclweb.org/portal/content/acl-code-ethics}).
We have taken measures to minimize unintended harm by carefully handling data and clearly communicating the limitations of our methodology.

\paragraph{Bias Propagation and Fairness.}
Despite efforts to mitigate biases by using multiple \acp{llm} and bidirectional comparisons, there remains a risk that patterns inherent in the underlying training data could be reproduced or even amplified. This may lead to systematic favoring or disadvantaging of specific types of content or writing styles. Users of this framework should remain aware of these inherent limitations and, where feasible, introduce additional safeguards—such as randomization, diverse model choices, or sampling checks—to reduce bias effects.

\paragraph{Over-Reliance on Automated Judgments, Accountability, and Transparency.}
Our methodology relies heavily on automated pairwise comparisons carried out by \acp{llm}. While this saves time and effort, it can overlook subtle contextual or domain-specific nuances that human experts might detect. Moreover, Elo ratings and numerical scores can create an illusion of objectivity and precision—potentially obscuring the subjective nature of \ac{llm} judgments, especially for high-stakes decisions. To address these concerns, we recommend maintaining a "human-in-the-loop" approach—where domain experts review and validate the automated scores to ensure they align with expert judgment. This could involve periodic audits or random sampling of the \ac{llm} outputs to ensure that the automated system is functioning as intended.

\paragraph{Privacy and Data Handling.}
Because the system repeatedly passes textual data (e.g., abstracts of research papers) to external \ac{llm} APIs, there is a risk of exposing sensitive or private information. In this work, we used only openly available data. Any private or proprietary data should be anonymized or stripped of identifying details prior to evaluation. Researchers must secure informed consent where necessary and ensure compliance with local data-protection regulations, as well as any terms of service imposed by LLM providers.

\paragraph{Responsible NLP Research Checklist.}
This paper addresses the relevant points outlined in the ACL Responsible NLP Research Checklist (\url{https://aclrollingreview.org/responsibleNLPresearch/}).

\textit{Checklist item A:}
The authors have added a dedicated limitations section, where we outline the methodological boundaries and constraints of our approach (see Section~\ref{sec:limitations}). In addition, we discuss potential risks and harms, including misuse and bias, in the Ethical Impact Statement (see Section~\ref{sec:ethical-impact}).

\textit{Checklist item B:}
An artifact in the form of code was created to demonstrate the proposed methodology and support reproducibility. No new data was collected for this work. The data used for evaluation purposes consisted of existing materials created by members of our institution and was used with appropriate permission.

\textit{Checklist item C:}
Yes, we conducted computational experiments using external \ac{llm} APIs, as described in the Implementation section (see Section~\ref{sec:implementation}). The experimental setup, including evaluation procedures and conditions, is detailed in the Evaluation section (see Section~\ref{sec:evaluation-strategy}).

\textit{Checklist item D:}
Yes, we involved human experts from our institution in the evaluation process. The experimental design and privacy considerations are described in detail in the Evaluation section (see Section~\ref{sec:evaluation-strategy}). All participants worked with their own publicly available abstracts, and no personal or sensitive data was collected.

\textit{Checklist item E:}
Yes, we used AI assistants during this research. ChatGPT was employed for language-related support, and GitHub Copilot assisted with coding. We describe the use of these tools in the Acknowledgments section (see Section~\ref{sec:acknowledgments}), while emphasizing that all authors retain full responsibility for the scientific content of the paper.

\bibliography{references_cleaned}

\appendix

\section{Appendix}
\label{sec:appendix}

\subsection{Large Language Models Used}
\label{sec:llms-used}

The following LLMs were integrated into the evaluation pipeline:
\begin{table}[ht]
  \centering
  \caption{LLMs used in the evaluation pipeline. Profile generation uses  temperature 0.5 with 6 completions; evaluation uses temperature 0,1. Quantization details are proprietary and undisclosed.}
  \label{tab:llm_parameters}
  \scriptsize
  \begin{tabular}{p{1.5cm}p{1cm}p{2.7cm}}
    \toprule
    \textbf{Model}   & \textbf{Role} & \textbf{API Access}                                                                 \\
    \midrule
    llama-3.1-70B    & Gen           & \href{https://console.groq.com/docs/models}{GROQ}: \texttt{llama-3.1-70b-versatile} \\
    \addlinespace[2pt]
    gemma2-9b-it     & Eval          & \href{https://console.groq.com/docs/models}{GROQ}: \texttt{gemma2-9b-it}            \\
    \addlinespace[2pt]
    llama-3.1-8b     & Eval          & \href{https://console.groq.com/docs/models}{GROQ}: \texttt{llama-3.1-8b-instant}    \\
    \addlinespace[2pt]
    gpt-4o-mini      & Eval          & \href{https://openai.com/api/}{OpenAI API}                                          \\
    \addlinespace[2pt]
    gemini-2.0-flash & Eval          & \href{https://ai.google.dev/gemini-api}{Google AI API}                              \\
    \addlinespace[2pt]
    mixtral-8x7b     & Eval          & \href{https://console.groq.com/docs/models}{GROQ}: \texttt{mixtral-8x7b-32768}      \\
    \bottomrule
  \end{tabular}
\end{table}

\subsection{Example Competency Profiles}
\label{example:competency-profiles}

The following two profiles outline the competencies of two experts in the field of information extraction and community development.

\begin{tcolorbox}[colback=gray!5!white, colframe=gray!75!black, title=Demonstrative Profile 1]

  \textbf{Domain Expertise:} Advancing information extraction through generative Large Language Models (LLMs)

  \medskip

  \textbf{Competencies:}
  \begin{itemize}
    \item \textbf{Information Extraction Technologies:} Utilizes generative LLMs for structural text analysis, identifying entities, relations, and events.

    \item \textbf{Cross-Domain Adaptability:} Applies LLMs across diverse domains, showcasing flexibility in understanding and generating domain-specific texts.

    \item \textbf{Systematic Literature Analysis:} Conducts in-depth reviews of contemporary research on LLM-based information extraction techniques.

    \item \textbf{Subtask-Based Taxonomy:} Categorizes advancements in LLM-driven information extraction by subtasks and underlying learning paradigms.

    \item \textbf{Trend Forecasting:} Identifies emerging trends and anticipates future directions in LLM applications for information extraction.

    \item \textbf{Community Contribution:} Curates and regularly updates a public repository of relevant research on LLM-enhanced information extraction.
  \end{itemize}

\end{tcolorbox}

\begin{tcolorbox}[colback=gray!5!white, colframe=gray!75!black, title=Demonstrative Profile 2]

  \textbf{Domain Expertise:} Facilitating the development of scientific web communities through detailed competence analysis.

  \medskip

  \textbf{Competencies:}
  \begin{itemize}
    \item \textbf{Competence Identification:} Extracts and delineates individual competences with precision based on scientific publication data.

    \item \textbf{Community Building:} Supports the formation and growth of research communities by aligning and harmonizing diverse areas of expertise.

    \item \textbf{Decision Support Systems:} Integrates structured competence data into advanced decision-making frameworks to enhance strategic outcomes.

    \item \textbf{Team Formation:} Enables effective team assembly through accurate competence mapping and role alignment.

    \item \textbf{Knowledge Visualization:} Employs sophisticated visualization tools to depict the development and interaction dynamics of virtual research communities.

    \item \textbf{Expertise Analysis:} Analyzes published research to identify optimal collaborations and recommend role assignments.
  \end{itemize}

\end{tcolorbox}

\subsection{Example Prompt}
\label{example:prompt}

The following listing presents the complete prompt structure used in our evaluation framework. The prompt demonstrates a multi-turn conversation between the system, user, and assistant, showcasing both an example evaluation and the actual task to be performed.

\begin{lstlisting}[
    caption={Complete Example Prompt for Competency Profile Evaluation},
    label={lst:complete-prompt}
]
(*@\rolelabel{systemmsg}{System}@*) You are a skilled evaluator tasked with evaluating the relevance of two competency profiles that were extracted by another system from provided scientific abstracts. Each profile is expected to reflect a specific domain of expertise and list 3 to at most 8 key competencies demonstrated by the author. Your task is to evaluate how well each profile reflects the competencies, themes, and expertise areas mentioned in the abstracts. Compare the two profiles and determine which one is more relevant to the abstracts, structuring your response as a JSON object as follows:
{
    "reasoning": "[Your Evaluation and Reasoning]",
    "preferred_profile": [1 or 2]
}
Your analysis should be neutral, accurate, and detailed, based on the content of the abstracts provided.

(*@\rolelabel{usermsg}{User}@*) Example 1:

Abstract 1:
Patients living in underserved areas do regularly express an interest in stone prevention; however, factors limiting participation, aside from obvious cost considerations, are largely unknown. To better understand factors associated with compliance with submitting 24-hour urine collections, we reviewed our patient experience at the kidney stone clinic at a hospital that provides care for an underserved urban community. A retrospective chart review of patients treated for kidney and/or ureteral stones between August 2014 and May 2016 was performed. Patient demographics, medical characteristics, stone factors, and compliance data were compiled into our data set. Patients were divided into two groups: those who did and did not submit the requested initial 24-hour urine collection. Analysis of factors related to compliance was performed using univariate analysis and multivariate logistic regression. A total of 193 patients met inclusion criteria for our study, 42.5% (82/193) of whom submitted 24-hour urine samples. Of the 82 collections submitted, 34.1% (28/82) were considered inadequate by creatinine level. A second urine collection within 6 months was obtained in 14.0% (27/193) of patients. Univariate analysis demonstrated that African American (AA) patients were less likely to submit an initial 24-hour urine collection than Caucasian patients (collected: 30.9% vs 51.8%; p < 0.05, respectively). Patients with a family history of kidney stones were more likely to submit an initial 24-hour urine collection than patients without a family history of kidney stones (61.1% vs 38.2%, p < 0.02, respectively). On multivariate analysis, both factors remained significant predictors of compliance with submitting a 24-hour urine collection. In our underserved patient population, AA patients were half as likely to submit a 24-hour urine collection than Caucasian patients, whereas patients with a positive family history of stones were more than twice as likely to submit than patients with no family history.

Abstract 2:
Iatrogenic ureteric injuries in gynecologic surgery are quite common. The laparoscopic spectra of treatment gives a wide range of application. We present the case of a 40-year-old female who underwent total abdominal hysterectomy with bilateral salpingo-oopherectomy for dysfunctional uterine bleeding. Postoperatively, she had oliguria with raised creatinine and was found to have ureteric narrowing on magnetic resonance urogram. The patient was successfully managed by laparoscopic (Boari flap) ureteroneocystostomy.

Profile 1:
Domain: "Healthcare and Patient Care"

Competencies:
- Patient Demographics Analysis: Demonstrated through the review of patient characteristics, including factors such as race and family history, to better understand factors associated with compliance in submitting 24-hour urine collections.
- Medical Chart Review: Showcased by conducting a retrospective chart review of patients treated for kidney and/or ureteral stones to compile relevant data for analysis.
- Statistical Analysis: Exhibited through the use of univariate analysis and multivariate logistic regression to identify factors related to compliance in submitting 24-hour urine collections.
- Clinical Research: Shown by investigating factors associated with compliance in submitting 24-hour urine collections in an underserved patient population.
- Patient Care: Displayed through the identification of factors that may limit patient participation in stone prevention, such as race and family history.
- Surgical Intervention: Demonstrated by presenting a case of a patient successfully managed by laparoscopic (Boari flap) ureteroneocystostomy following iatrogenic ureteric injuries in gynecologic surgery.
- Diagnostic Imaging Interpretation: Showcased through the interpretation of magnetic resonance urogram to identify ureteric narrowing in a patient following gynecologic surgery.
- Clinical Practice Improvement: Exhibited by identifying actionable insights, such as the difference in compliance between African American and Caucasian patients, to improve patient care and outcomes in underserved populations.

Profile 2:
Domain: "Urology and Surgical Care"

Competencies:
- Patient Data Analysis: Demonstrated through the retrospective chart review and analysis of patient demographics, medical characteristics, and compliance data to understand factors associated with patient participation in medical studies.
- Surgical Procedure Expertise: Showcased through the successful management of a patient with iatrogenic ureteric injury using laparoscopic ureteroneocystostomy, specifically the Boari flap technique.
- Research Study Design: Exhibited by designing and conducting a study to investigate factors related to patient compliance with submitting 24-hour urine collections, including data collection and statistical analysis.
- Statistical Analysis: Demonstrated through the use of univariate analysis and multivariate logistic regression to identify significant predictors of patient compliance.
- Cultural Competence: Shown by recognizing and addressing disparities in patient compliance among different demographic groups, such as African American patients being less likely to submit 24-hour urine collections.
- Medical Knowledge: Exhibited by understanding the clinical significance of creatinine levels, family history of kidney stones, and other medical factors that influence patient compliance and treatment outcomes.
- Case Management: Demonstrated by effectively managing a patient's postoperative complications, including oliguria and raised creatinine, and providing appropriate treatment and follow-up care.
- Clinical Problem-Solving: Showcased by identifying and addressing complex medical issues, such as iatrogenic ureteric injuries, and developing effective treatment plans to improve patient outcomes.

(*@\rolelabel{assistantmsg}{Assistant}@*) {
    "reasoning": " Both competency profiles demonstrate relevance to the provided scientific abstracts, focusing on healthcare, patient care, and urology. However, Profile 1 is more comprehensive in covering the breadth of skills and competencies reflected in the abstracts, including patient demographics analysis, medical chart review, statistical analysis, clinical research, patient care, surgical intervention, diagnostic imaging interpretation, and clinical practice improvement. Profile 1 explicitly mentions the use of statistical methods (univariate analysis and multivariate logistic regression) and highlights the importance of understanding patient demographics and clinical characteristics in improving patient care and outcomes. Additionally, it covers both the clinical and research aspects presented in the abstracts, such as the management of patients with kidney stones and the investigation of factors influencing compliance with medical procedures. Profile 2, while also relevant, seems to emphasize more the surgical and research design aspects, without fully encapsulating the range of competencies demonstrated in the abstracts, such as the detailed analysis of patient demographics and the application of diagnostic imaging. Therefore, Profile 1 is more aligned with the competencies and themes presented in the abstracts.",
    "preferred_profile": 1
}

(*@\rolelabel{usermsg}{User}@*) Please assess the following competency profile in terms of its relevance to these scientific abstracts.

Abstracts:
Abstract 1:
In Computer Science, Software Engineering, Business Informatics or Information Systems, conceptual modeling is an important tool and as such also contained in the respective curricular recommendations. Especially in large university courses, an automated assessment of models can improve the quality of teaching and learning. While there are many different approaches to automatically assess conceptual models, these approaches, however, often only tackle a single aspect or a single type of conceptual model. In this paper, we aim to take a comprehensive perspective on the topic and shed light on the current state of the art and technique. Furthermore, as assessment approaches have to be developed in accordance with appropriate teaching or learning activities and desired learning outcomes, we inquire in which settings automated assessment approaches are included and to which extent didactic aspects are taken into account. To this end, we have conducted a systematic literature review in which we identified 110 relevant publications on the topic which we have analyzed in a structured way. The results provide answers to five relevant research questions and pinpoint open issues which should be inquired in further research.

Abstract 2:
In vielen Anwendungsbereichen der Informatik spielt die grafische Modellierung eine wichtige Rolle. Grafische Modelle kommen beispielsweise bei der Gesch\"aftsprozessmodellierung oder im Rahmen der Softwareentwicklung zum Einsatz, um komplexe Sachverhalte \"ubersichtlich darzustellen. In der Hochschullehre kommt derzeit eine kompetenzorientierte Ausrichtung entsprechender Lehrveranstaltungen zu kurz, ebenso sind die M\"oglichkeiten zur technischen Unterst\"utzung eingeschr\"ankt. Die in dieser Arbeit behandelten Forschungsfragen sind daher einer kompetenzorientierten Ausrichtung des Pr\"ufens auf dem Gebiet der grafischen Modellierung sowie der Entwicklung einer entsprechenden E-Assessment-Plattform gewidmet. Im Rahmen der Arbeit wurde anhand theoriebasierter und empirischer Ans\"atze ein umfassendes Kompetenzmodell entwickelt, das Lernziele f\"ur zentrale Handlungsbereiche der grafischen Modellierung und \"uberfachliche Kompetenzen beschreibt. Es wurde ein Aufgabenkatalog erstellt, der Aufgabentypen mit den im Kompetenzmodell definierten Lernzielen verkn\"upft. Erg\"anzend wurden exemplarische Bewertungsschemata und Empfehlungen f\"ur die Gestaltung lernf\"orderlicher Feedback-Nachrichten auf Basis des Kompetenzmodells abgeleitet. Die Ergebnisse unterst\"utzen Lehrende bei der Auswahl von Lernzielen und der Gestaltung kompetenzorientierter Pr\"ufungen anhand passender Modellierungsaufgaben. Zur Umsetzung kompetenzorientierter Pr\"ufungen auf dem Gebiet der grafischen Modellierung wurde eine E-Assessment-Plattform entwickelt. Diese ber\"ucksichtigt verschiedene grafische Modellierungssprachen, individuelle Bewertungsschemata und Feedback-Empfehlungen. Zus\"atzlich wurden Dienste zur automatisierten Bewertung von Petri-Netzen erstellt, die Lernziele zu syntaktischen, semantischen und pragmatischen Qualit\"atsaspekten adressieren. Die Einsatzf\"ahigkeit der Plattform wurde im praktischen Einsatz in Lehrveranstaltungen und Pr\"ufungen demonstriert. Erg\"anzend wurden Befragungen zur Benutzungsfreundlichkeit und weiteren Aspekten durchgef\"uhrt und die Ergebnisse der Anwendung der Bewertungsdienste auf einer umfangreichen Datenbasis studentischer Petri-Netze evaluiert.

Abstract 3:
Using e-learning and e-assessment environments in higher education bears considerable potential for both students and teachers. In this contribution we present an architecture for a comprehensive e-assessment platform for the modeling domain. The platform -- currently developed in the KEA-Mod project -- features a micro-service architecture and is based on different inter-operable components. Based on this idea, the KEA-Mod platform will provide e-assessment capabilities for various graph-based modeling languages such as Unified Modeling Language (UML), EntityRelationship diagrams (ERD), Petri Nets, Event-driven Process Chains (EPC) and the Business Process Model and Notation (BPMN) and their respective diagram types.

Abstract 4:
In vielen Bereichen der Wirtschaftsinformatik spielt die Erstellung konzeptueller Modelle unter Verwendung grafischer Modellierungssprachen eine wichtige Rolle. Entsprechend wichtig ist eine fundierte Grundausbildung, die sich an den ben\"otigten Modellierungskompetenzen orientiert und daher neben theoretischen auch praktische Aspekte der konzeptuellen Modellierung in den Blick nimmt. Der vorliegende Beitrag stellt erste Ergebnisse aus dem KEA-Mod-Projekt vor, das sich mit der Erstellung eines "digitalen Fachkonzepts" im Bereich der grafischen, konzeptuellen Modellierung befasst. Kernst\"uck dieses Fachkonzepts ist die Unterst\"utzung der Grundausbildung in der grafischen, konzeptuellen Modellierung durch eine kompetenzorientierte E-Assessment-Plattform mit automatisierten und individuellen Bewertungs- und Feedbackm\"oglichkeiten.

Abstract 5:
Die KEA-Mod-Plattform erm\"oglicht es, Modellierungsaufgaben mit verschiedenen Modellierungssprachen wie z.B. UML, Petri-Netzen, EPK oder BPMN durch Dozierende zu erstellen und von Studierenden bearbeiten zu lassen. Die Plattform kam in einer gro\"sen Lehrveranstaltung mit ca. 250 Studierenden zum Piloteinsatz. Die Studierenden konnten mit Hilfe der Plattform und des integrierten Modellierungswerkzeugs eine Aufgabenreihe mit Modellierungsaufgaben zu Petri-Netzen bearbeiten und einreichen. Anschlie\"send erhielten die Studierenden automatisiert generiertes Feedback. Das Poster beschreibt die Evaluation dieses Piloteinsatzes aus der Perspektive der Studierenden und bietet erste Ergebnisse in Bezug auf die Plattform-Usability und zur wahrgenommenen Lernf\"orderlichkeit des Feedbacks.

Profile 1:
Domain: "Graphical Modeling"

Competencies:
- Automated Assessment of Conceptual Models: Demonstrated across Abstracts 1, 2, and 5, this competency involves the development of automated assessment tools to evaluate conceptual models in various educational settings.
- Understanding of Different Approaches to Conceptual Modeling: Found in Abstracts 1 and 2, this competency involves an understanding of various approaches to conceptual modeling, including automated assessment techniques.
- Development of a Comprehensive E-Assessment Platform: Demonstrated across Abstracts 3 and 4, this competency involves the development of a comprehensive e-assessment platform for the modeling domain, incorporating different inter-operable components.
- Understanding of Different Graph-Based Modeling Languages: Found in Abstracts 3 and 4, this competency involves an understanding of different graph-based modeling languages, including UML, Petri Nets, and BPMN.
- Development of a Micro-Service Architecture: Demonstrated across Abstracts 3 and 4, this competency involves the development of a micro-service architecture for the e-assessment platform.
- Understanding of the Importance of Feedback in Learning: Found in Abstracts 3 and 4, this competency involves an understanding of the importance of feedback in learning, particularly in the context of graphical modeling.
- Evaluation of the Effectiveness of Automated Assessment Tools: Demonstrated across Abstracts 1, 2, 3, and 4, this competency involves the evaluation of the effectiveness of automated assessment tools in various educational settings.
- Understanding of the Role of Graphical Modeling in Teaching and Learning: Found in Abstracts 1 and 2, this competency involves an understanding of the role of graphical modeling in teaching and learning, particularly in the context of conceptual modeling.

Profile 2:
Domain: "Graphical Modeling"

Competencies:
- Automated Assessment: Has experience with automated tools for evaluating models, but may not fully understand the nuances of different assessment approaches.
- Conceptual Modeling: Familiar with various modeling techniques, but lacks depth in understanding the theoretical foundations and differences between approaches.
- E-Assessment Platform Development: Involved in the development of an e-assessment platform, but may not have a comprehensive understanding of its architecture or the integration of different components.
- Graph-Based Modeling Languages: Knows about several graph-based modeling languages, including UML and Petri Nets, but may not be proficient in their application or the nuances of each.
- Micro-Service Architecture: Aware of the concept of micro-service architecture, but may not fully grasp its implementation or benefits in the context of the e-assessment platform.
- Feedback in Learning: Recognizes the importance of feedback, but may not have a deep understanding of how to effectively implement feedback mechanisms in graphical modeling education.
- Evaluation of Assessment Tools: Has some experience with evaluating the effectiveness of automated assessment tools, but may not have a systematic approach to this process.
- Role of Graphical Modeling: Understands that graphical modeling is used in teaching and learning, but may not fully appreciate its significance or how it contributes to student learning outcomes.

Your evaluation must follow this JSON format:
{
    "reasoning": "[Your Reasoning and Evaluation in that order]",
    "preferred_profile": [1 or 2]
}
Be specific and detailed in your reasoning and provide the number of the preferred profile.

(*@\rolelabel{assistantmsg}{Assistant}@*) {
    "reasoning": "Profile 1 is significantly more relevant to the provided scientific abstracts than Profile 2. It comprehensively captures the competencies demonstrated in the abstracts, such as the development of automated assessment tools for conceptual models, the understanding of various graph-based modeling languages, and the creation of a comprehensive e-assessment platform. Profile 1 also emphasizes the importance of feedback in learning and the evaluation of automated assessment tools, which are central themes in the abstracts. In contrast, Profile 2 presents a more superficial understanding of these competencies, indicating familiarity without demonstrating depth or comprehensive knowledge. For instance, while it mentions automated assessment and e-assessment platform development, it lacks the detailed insights and systematic approaches that Profile 1 provides. Additionally, Profile 1 effectively connects the competencies to specific aspects of the abstracts, such as the systematic literature review and the development of a competence model, which are crucial for understanding the context and significance of the research. Therefore, Profile 1 is the preferred choice as it aligns more closely with the themes and expertise areas highlighted in the abstracts.",
    "preferred_profile": 1
}
\end{lstlisting}

\end{document}